\documentclass[runningheads]{llncs}
\usepackage{graphicx}
\usepackage{amsmath,amssymb} % define this before the line numbering.
\usepackage{color}
\usepackage{cite}
\usepackage{ragged2e}
\usepackage{bm}
\usepackage{subcaption}
\usepackage{cleveref}
\usepackage{booktabs}
\usepackage{pgfplots}
\usepackage{scalefnt}

\begin{document}
\def\agl#1{\langle#1\rangle}
\def\Dataset#1{\mathcal{D}_#1}
\def\baseData{\Dataset{B}}
\def\novelData{\Dataset{V}}
\def\extractor#1{f_{\theta}(#1)}
\def\classN{N}
\def\imageN{K}
\def\queryN{Q}
\def\subSet#1{\mathcal{#1}}
\def\supportSet{\subSet{S}}
\def\querySet{\subSet{Q}}
\def\task{\mathcal{T}=\{\left(\supportSet, \querySet\right)\}}
\def\image#1{x_{#1}}
\def\lb#1{y_{#1}}
\def\datapair#1{(\image{#1}, \lb{#1})}
\def\supportDe{\supportSet=\{\datapair{i}\}_{i=1}^{\imageN\times\classN}}
\def\distance#1#2{d({#1},{#2})}
\def\proto#1{P_{#1}}

\def\gImage{I}
\def\mean#1{\boldsymbol{\mu_{#1}}}
\def\var#1{\boldsymbol{\Sigma_{#1}}}
\def\guassdis#1{N_{#1}(\mean{#1},\var{#1}^2)}
\def\gpatch#1{{p}_{#1}}
\def\gpatchN{M}
\def\featuredim{D}
\def\err#1{\boldsymbol{e_{#1}}}
\def\feature{\boldsymbol{f}}

\definecolor{color0}{RGB}{12,33,134}
\definecolor{color1}{RGB}{145,30,180}
\definecolor{color2}{RGB}{245,130,48}
\definecolor{color3}{RGB}{230,25,75}

\title{Gestalt-Guided Image Understanding for Few-Shot Learning}
\author{Kun Song \and
Yuchen Wu \and
Jiansheng Chen \and
Tianyu Hu \and
Huimin Ma\thanks{Corresponding author.}}
\authorrunning{K. Song et al.}
\institute{University of Science and Technology Beijing, Beijing, China \\
\email{\{songkun,yuchen.wu\}@xs.ustb.edu.cn \\ \{jschen,Tianyu,mhmpub\}@ustb.edu.cn }}
\maketitle              % typeset the header of the contribution
\begin{abstract}
Due to the scarcity of available data, deep learning does not perform well on few-shot learning tasks. However, human can quickly learn the feature of a new category from very few samples. Nevertheless, previous work has rarely considered how to mimic human cognitive behavior and apply it to few-shot learning. This paper introduces Gestalt psychology to few-shot learning and proposes Gestalt-Guided Image Understanding, a plug-and-play method called \textbf{GGIU}. Referring to the principle of totality and the law of closure in Gestalt psychology, we design Totality-Guided Image Understanding and Closure-Guided Image Understanding to extract image features. After that, a feature estimation module is used to estimate the accurate features of images. Extensive experiments demonstrate that our method can improve the performance of existing models effectively and flexibly without retraining or fine-tuning. Our code is released on \url{https://github.com/skingorz/GGIU}.

% \keywords{First keyword  \and Second keyword \and Another keyword.}
\end{abstract}
\section{Introduction}

In recent years, deep learning has shown surprising performance in various fields. Nevertheless, deep learning often relies on large amounts of training data. More and more pre-trained models are based on large-scale data. For example, CLIP \cite{DBLP:conf/icml/RadfordKHRGASAM21} is trained on 400 million image-text pairs. However, a large amount of data come with extra costs in deep learning procedures, such as collection, annotation, and training. In addition, many kinds of data, such as medical image data, requires specialized knowledge to annotate. Data for some rare scenes are hard to obtain, such as car accidents. Therefore, there is a growing interest in training a better model using fewer data. Motivated by this, Few-Shot Learning (FSL)\cite{DBLP:journals/pami/Fei-FeiFP06, DBLP:conf/nips/VinyalsBLKW16} is proposed to solve the problem of learning from small amounts of data.

The most significant obstacle to few-shot learning is the lack of data. In order to address this obstacle, existing few-shot learning approaches mainly employ metric learning, such as PN\cite{DBLP:conf/nips/SnellSZ17}, and meta-learning, such as MAML\cite{DBLP:conf/icml/FinnAL17}. Regardless of the technique, the ultimate goal is to extract more robust features for novel classes. Previous researches mainly focus on two aspects: designing a more robust feature extractor to represent the image feature better, such as meta-baseline\cite{DBLP:conf/iccv/Chen00D021}, and using more dense features, such as DeepEMD\cite{DBLP:conf/cvpr/ZhangCLS20}. However, few people consider how to mimic human learning patterns to enhance the effectiveness of few-shot learning.

\begin{figure}[tbp]
	\centering
	\begin{subfigure}[t]{0.45\textwidth}
		\centering
		\includegraphics[width=0.9\linewidth]{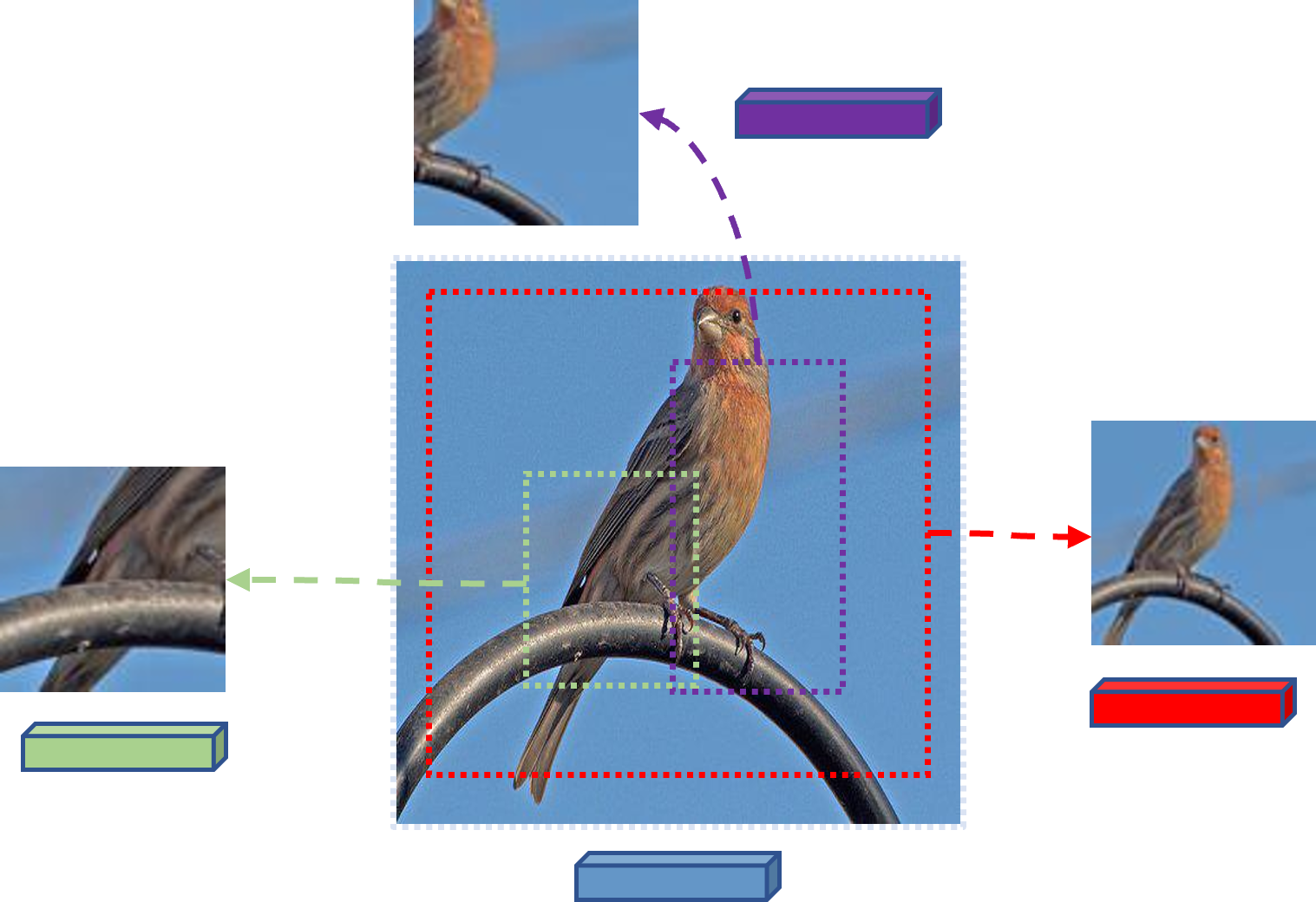}
% 		\caption{A bird image and some patches randomly cropped from the image.}
		\label{birds}
	\end{subfigure}
	\hfill
	\begin{subfigure}[t]{0.45\textwidth}
		\centering
		\includegraphics[width=0.9\linewidth]{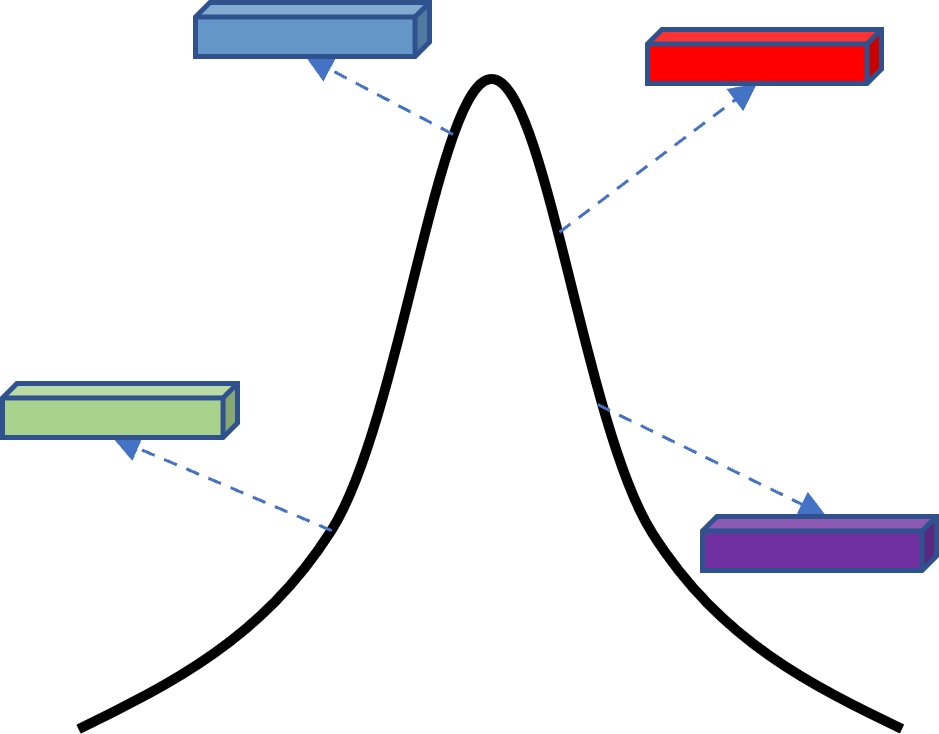}
        % \includesvg[width=0.5\linewidth]{image/gauss.svg}
% 		\caption{The potential distribution of the image.}
		\label{gauss}
	\end{subfigure}
	\caption{Describe an image using a multivariate Gaussian distribution.}
	\label{birdimage}
\end{figure}

Modern psychology has extensively studied the mechanisms of human cognition. Gestalt psychology is one of them. On the one hand, Gestalt psychology states that conscious experience must be considered globally, which is the principle of totality of gestaltism. For example, as shown in the left of \textbf{Fig. \ref{birdimage}}, the whole picture shows a bird standing on a tree branch. Its feature can represent the image. Meanwhile, given any patch of the image, human can easily determine that it is a part of a bird. These can be explained by the law of closure in Gestalt psychology: When parts of a whole picture are missing, our perception fills in the visual gap. Therefore, the feature of patches can also represent the image.

Motivated by Gestalt psychology, we imitate human learning patterns to redesign the image understanding process and apply it to few-shot learning. In this paper, we innovatively describe the image as a data distribution to represent the principle of totality and the law of closure in Gestalt psychology. We assume that image can be represented by a  corresponding multivariate Gaussian distribution. As shown in the right of \textbf{Fig.\ref{birdimage}},
the feature of the image can be considered as a sample of the potential distribution. Likewise, the feature of the largest patch, the image itself, is also a sample of the distribution. Then we design a feature estimation module with reference to Kalman filter to estimate the features of images accurately.

The main contributions of this paper are:

% 我们将格式塔理论引入到图像理解过程中，提出了一种即插即用无需微调的方法，GGIU.

1.We introduce Gestalt psychology into the process of image understanding and propose a plug-and-play method without retraining or fine-tuning, called \textbf{GGIU}.

2. We innovatively propose to use multivariate Gaussian distribution to describe the image and design a feature estimation module with reference to Kalman filter to estimate image feature accurately.

3. We conduct extensive experiments to demonstrate the applicability of our method to a variety of different few-shot classification tasks. The experiment results demonstrate the robustness and scalability of our method.

\section{Related Work}
\label{sec:relatedwork}

\subsection{Few-shot Learning}
\label{related:fsl}

Most of the existing methods for few-shot learning is based on the meta-learning \cite{DBLP:journals/corr/abs-2004-05439} framework. The motivation of meta-learning is learning to learn \cite{DBLP:books/sp/98/ThrunP98}. The network is trained on a set of meta-tasks during the training process to gain the ability to adapt itself to different tasks. The meta-learning methods for few-shot learning are mainly divided into two categories, optimization-based and metric-based.

 Koch \cite{koch2015siamese} apply metric learning to few-shot learning for the first time. They proposed to apply Siamese Neural Network to one-shot learning. A pair of convolutional neural networks with shared weights is used to extract the embedding features of each class separately. When inferring the category of unknown data, the unlabeled data and the training set samples are paired. The Manhattan Distance between unlabeled data and training data is calculated as the similarity, and the category with the highest similarity is used as the prediction of the samples. Matching Network\cite{DBLP:conf/nips/VinyalsBLKW16} first conducts experiments on \emph{mini}ImageNet for few-shot learning. It proposes an attention module that uses cosine distance to determine the similarity between the target object and each category and uses the similarity for the final classification. Prototypical Network\cite{DBLP:conf/nips/SnellSZ17} proposed the concept of category prototypes. Prototypical Network takes a class’s prototype to be the mean of its support set in the embedding space. The similarity between the unknown data and each category's prototypes are measured, and the most similar category is selected as the final classification result. Satorras\cite{DBLP:conf/iclr/SatorrasE18} uses graphical convolutional networks to transfer information between support and query sets and extended prototypical networks and matching networks to non-euclidean spaces to assist few-shot learning. DeepEmd \cite{DBLP:conf/cvpr/ZhangCLS20} adopt the Earth Mover’s Distance (EMD) as a metric to compute a structural distance between dense image representations to determine image relevance. COSOC \cite{DBLP:conf/nips/LuoWWYXXT21} extracts image foregrounds using contrast learning to optimize the category prototypes. Yang\cite{DBLP:conf/iclr/YangLX21} takes the perspective of distribution estimation to rectify the category prototype. SNE \cite{DBLP:conf/ijcai/TangTZ021} encodes the latent distribution transferring from the already-known classes to the novel classes by label propagation and self-supervised learning. CSS \cite{DBLP:conf/ijcai/AnXZZ21} propose conditional self-supervised learning with a 3-stage training pipeline. CSEI\cite{li2021learning} proposed an Erasing-then-Inpainting method to augment the data while training, which needs to retrain the model. AA\cite{afrasiyabi2020associative} expands the novel data by adding extra ``related base " data to few novel ones and fine-tunes the model.
 
 \begin{figure}[tp]
    \centering
    \includegraphics[width=.5\linewidth]{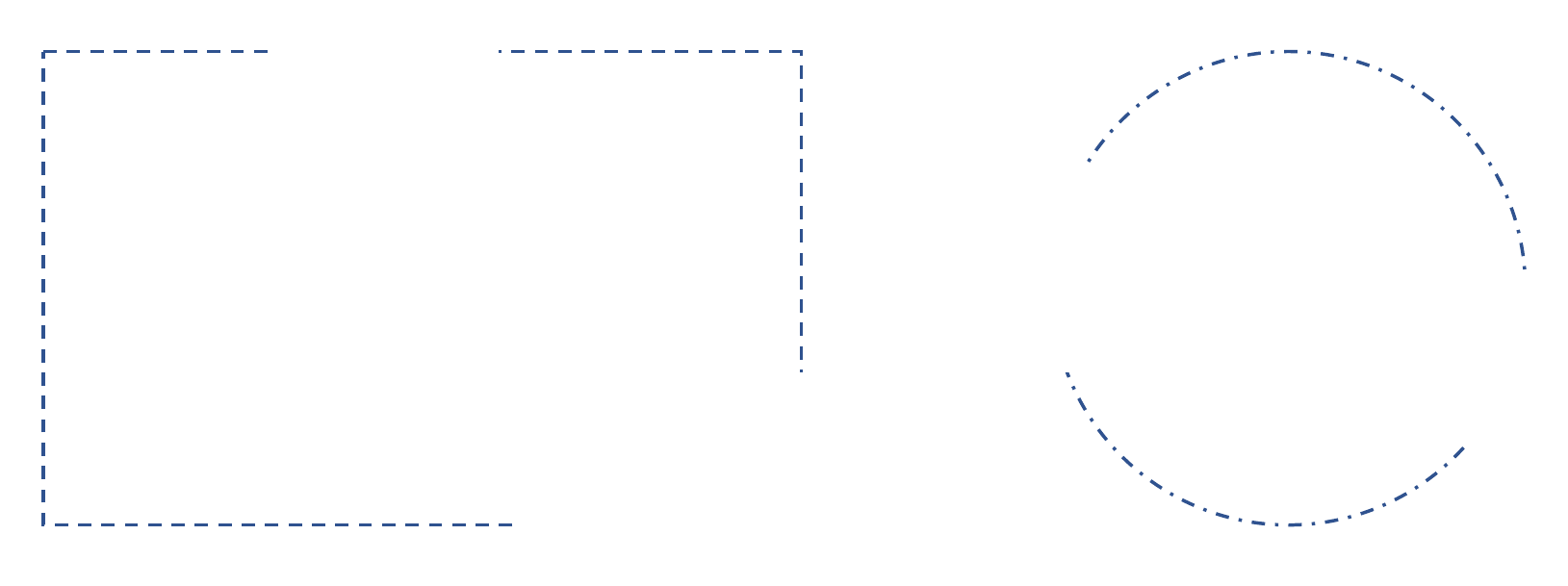}
    % \includesvg[width=\linewidth]{image/method.svg}
    \caption{Although we see incomplete figures, our brain can easily complete them and regard them as rectangles (left) and circles (right).}
    \label{fig:closure}
\end{figure}

\subsection{Gestalt Psychology}
\label{related:Gestalt}

Gestalt psychology is a psychology school that emerged in Austria and Germany in the early twentieth century. Gestalt principles, proximity, similarity, figure-ground, continuity, closure, and connection describe how humans perceive visuals in connection with different objects and environments. In this section, we mainly introduce the principle of totality and  the law of closure.

\subsubsection{Principle of Totality} 

% The principle of totality points out that conscious experience must be considered globally. Wertheimer \cite{wagemans2012century} described holism as fundamental to Gestalt psychology: ``A perceptual whole is different from what one would predict based on only its individual parts." Moreover, the nature of a part depends upon the whole in which it is embedded. Kohler\cite{henle1972selected} thinks: ``In psychology, instead of being the sum of parts existing independently, wholes give their parts specific functions or properties that can only be defined in relation to the whole in question." Thus, the maxim that the whole is more than the sum of its parts is not a precise description of the Gestaltist view. \cite{wagemans2012century} Instead, ``The whole is something else than the sum of its parts, because summing is a meaningless procedure, whereas the whole-part relationship is meaningful."\cite{koffka2013principles}

The principle of totality points out that conscious experience must be considered globally. Wertheimer \cite{wagemans2012century} described holism as fundamental to Gestalt psychology. Kohler \cite{henle1972selected} thinks: “In psychology, instead of being the sum of parts existing independently, wholes give their parts specific functions or properties that can only be defined in relation to the whole in question.” Thus, the maxim that the whole is more than the sum of its parts is not a precise description of the Gestaltist. \cite{wagemans2012century}

\subsubsection{Law of Closure} Gestalt psychologists held the view that humans often consider objects as complete rather than focusing on the gaps they may have.\cite{hamlyn2017psychology} 
For example, a circle has a good Gestalt in terms of completeness. However, we may also consider an incomplete circle as a complete one. This tendency to complete shapes and figures is called closure. \cite{brennan2017history} The law of closure states that even incomplete objects, such as forms, characters, and pictures, are seen as complete by people. In particular, when part of the entire picture is missing, our perception fills in the visual gaps. For example, as shown in \textbf{Fig. \ref{fig:closure}}, despite the incomplete shape, we still perceive a rectangle and a circle. If the law of closure did not exist, the image would depict different lines with different lengths, rotations, and curvatures. However, because of the law of closure, we perceptually combine the lines into whole shapes.\cite{stevenson2012emergence}

\section{Method}
\label{sec:method}

\subsection{Problem Definition}
\label{sec:problemDefinition}
The dataset for few-shot learning consists of training set $\baseData$ and testing set $\novelData$ with no shared classes. We train a feature extractor $\extractor{\cdot}$ on $\baseData$ containing lots of labeled data. During the evaluation, many $\classN$-way $\imageN$-shot $\queryN$-query tasks $\task$ are constructed from $\novelData$. Each task contains a support set $\supportDe$ and a query set $\querySet$. Firstly, $\classN$ classes in $\novelData$ are randomly sampled for $\supportSet$ and $\querySet$. Then we calculate a classifier for $\classN$-way classification for each task based on $\extractor{\cdot}$ and $\supportSet$. At last, we calculate the feature of the query image $\image{} \in \querySet$ and classify it into one of the $\classN$ class.

\subsection{Metric-based Few-shot Learning Pipeline}
\label{sec:pipeline}

Calculating a better representation for all classes in metric-based few-shot learning is critical. Under most circumstances, the features of all images in each category of the support set are calculated separately. The mean of features is used as the category representation, the prototype.

\begin{equation}
    \label{equ:proto}
    \boldsymbol{\proto{n}}=\frac{1}{\imageN{}}\sum_{\datapair{i}\in\supportSet_n}\extractor{\image{i}}
\end{equation}

In \textbf{Equation \ref{equ:proto}}, $\boldsymbol{\proto{n}}$ represents the prototype of the $n$-th category and $\supportSet_n$ represents the set of all images of the $n$-th class. When an image $\image{} \in \querySet{}$ and a distance function $\distance{}{}$ are given, the feature extractor $\extractor{\cdot}$ is used to calculate the feature of $\image{}$. Then we calculate the distance between $\extractor{\image{}}$ and the prototypes in the embedding space. After that, a softmax function(\textbf{Equation~\ref{equ:softmax}}) is used to calculate a distribution for classification.

% Todo check this equation
\begin{equation}
    \label{equ:softmax}
    p_\theta(y=n|x)=\frac{exp(-\distance{\extractor{x}}{{\boldsymbol{\proto{n}}}})}
    {\sum_{i=1}^{\classN{}}exp(-\distance{\extractor{x}}{{\boldsymbol{\proto{i}}}})}
\end{equation}

Finally, the model is optimized by minimizing the negative log-likelihood $L(\theta)=-\log p_\theta(y=n|x)$, i.e.,

\begin{equation}
    \label{neglogpro}
    % L(\theta)=-log(\frac{exp(-\distance{\extractor{x}}{{\proto{n}}})}
    % {\sum_{i=1}^{k}exp(-\distance{\extractor{x}}{{\proto{i}}})})
    NLL(\theta)=\distance{\extractor{x}}{{\boldsymbol{\proto{n}}}}+\log\sum_{i=1}^{k}exp(-\distance{\extractor{x}}{{\boldsymbol{\proto{i}}}})
\end{equation}

\subsection{Gestalt Guide Image Understanding}
Previous works mainly calculate prototypes with the features of images. Nevertheless, due to the limit of data volume, the image feature does not represent the information of the class well. In this section, inspired by Gestalt psychology, we reconceptualize images in terms of the principle of totality and the law of closure separately.

Given one image $\gImage{}$, many patches of different sizes can be cropped from it. As shown in the right of \textbf{Fig.\ref{birdimage}}, we assume that all patches follow a potential multivariate Gaussian distribution. All the patches can be regarded as samples of this distribution. Likewise, the largest patch of this image, the image itself, is also a sample of the distribution. Therefore, this potential Gaussian distribution can describe the image. The above can be expressed as follows: given a feature extractor $\extractor{\cdot}$, for any patch $\gpatch{}$ cropped from image $\gImage{}$, its feature follows a multivariate Gaussian distribution, i.e., $\extractor{\gpatch{}} \in \mathbb{R}^D$ and $ \extractor{\gpatch{}} \sim \guassdis{\gImage{}}$. Next, $\mean{I}$ can represent the feature of $I$. Finally, we estimate the image feature by the principle of totality and the law of closure.

\subsubsection{Totality-Guided Image Understanding}
\label{sec:totality}

The existing image understanding processes in few-shot learning are most from the totality of the image.  The image can be considered as a sample of the potential multivariate Gaussian distribution, i.e., $\extractor{\gImage{}}\sim\guassdis{\gImage{}}$. Therefore, $\guassdis{\gImage{}}$ can be estimated by $\extractor{\gImage{}}$ (\textbf{Equation \ref{estimateN1}}). $\mean{t}$ represents the estimate guided by the principle of totality.

\begin{equation}
\label{estimateN1}
    \mean{t}=\hat{\mean{\gImage{}}}=\extractor{\gImage{}}
\end{equation}

\subsubsection{Closure-Guided Image Understanding}
\label{sec:closure}

Guided by the law of closure, we randomly crop patches from images as the sample of the potential distribution of image. For any patch $\gpatch{}\in\gImage{}$, $\extractor{p}\in\mathbb{R}^\featuredim{}$, $\extractor{\gpatch{}} \sim \guassdis{I}$. 
The joint probability density function of the feature of the $i$-th patch is shown as \textbf{Equation~\ref{distr}}.

\begin{equation}
    \label{distr}
    p_{N_I}
    (\extractor{\gpatch{i}})=
    \frac{1}{(2\pi)^{\frac{\featuredim{}}{2}}|\var{}|^{-\frac{1}{2}}}
    \exp{\left(-\frac{1}{2}\left(\extractor{\gpatch{i}}-\mean{\gImage{}}\right)^T\var{}^{-1}\left(\extractor{\gpatch{i}}-\mean{\gImage{}}\right)\right)}
\end{equation}

The log-likelihood function is:

\begin{equation}
\begin{split}
    \ell_{N_I}(\mean{I},\var{I};\gpatch{1},\ldots,\gpatch{M})= 
    & -\frac{\gpatchN{}\featuredim{}}{2}log(2\pi)-\frac{\gpatchN{}}{2}log(|\var{I}|) \\
    & -\frac{1}{2}\sum\limits_{j=1}\limits^{\gpatchN{}}(\gpatch{i}-\mean{I})^T{\var{I}^{-1}}(\gpatch{i}-\mean{I})
    \end{split}
\end{equation}

Solve the following maximization problem

\begin{equation}
    \label{equ:maxlll}
    \max\limits_{\mean{I},\var{I}} \ell_{N_I}(\mean{I},\var{I};\gpatch{1},\ldots,\gpatch{M}) 
\end{equation}

Then we have

\begin{equation}
\label{estimateN2}
    \mean{c}=\hat{\mean{I}}=\frac{1}{\gpatchN{}}\sum_{i=1}^\gpatchN{}\extractor{\gpatch{i}}
\end{equation}

$\mean{c}$ represents the estimation guided by the law of closure.

\subsection{Feature Estimation}
\label{sec:lambda}

We regard the process of estimating image feature guided by the totality and closure as two different observers following multivariate Gaussian distribution: $O_t$ and $O_c$.  The observations of $O_t$, $O_c$ are $\mean{t} \in \mathbb{R}^{D\times 1}$, $\mean{c} \in \mathbb{R}^{D\times 1}$ and their random errors are $\err{t}$, $\err{c}$ respectively, where $\err{t} \in \mathbb{R}^{D\times 1}$ and $\err{c} \in \mathbb{R}^{D\times 1}$. $\err{t}$ and $\err{c}$ follow multivariate Gaussian distribution, i.e., $\err{t} \sim {N(\mathbf{0},\var{t}^2)}$,  $\err{c}~\sim~{N(\mathbf{0},\var{c}^2)}$, where $\var{t} \in \mathbb{R}^{D\times D}$ and $\var{c} \in \mathbb{R}^{D\times D}$. In this section, we use Kalman filter to estimate image features $\feature{} \in \mathbb{R}^{D\times 1}$. 

For $O_t$ and $O_c$, we have
\begin{align}
    & \feature{} = \mean{t} + \err{t}  \\
    & \mean{c} =  \feature{} + \err{c} 
\end{align}

The prior estimates of $\feature{}$ under $O_t$ and $O_c$ are
\begin{align}
    & \hat{\feature{}^-_t} = \mean{t}   \\
    & \hat{\feature{}^-_c} = \mean{c}  
\end{align}

The feature $\feature{}$ can be estimate by the prior estimates under $O_t$ and $O_c$, we have
\begin{equation}
    % \hat{\feature{}} = \hat{\feature{}^-_t} + \boldsymbol{\lambda}(\hat{\feature{}^-_c}-\hat{\feature{}^-_t}) 
    \hat{\feature{}} = \hat{\feature{}^-_c} + \boldsymbol{\lambda}(\hat{\feature{}^-_t}-\hat{\feature{}^-_c}) 
\end{equation}

i.e.,
\begin{equation}
\label{equ:imagefea}
    \hat{\feature{}}=\mean{c} + \boldsymbol{\lambda}(\mean{t}-\mean{c})
\end{equation}

$\boldsymbol{\lambda}=diag(\lambda_1,\lambda_2,\ldots,\lambda_D)$, where $\lambda_i$ is a diagonal matrix, ranging from 0 to $\boldsymbol{I}$. The error between $\hat{\feature{}}$ and $\feature{}$ is
\begin{equation}
    \err{}=\feature{} - \hat{\feature{}} 
\end{equation}

The error $\err{}$ follows a multivariate Gaussian distribution, i.e., $\err{} \sim{} N(\boldsymbol{0},\var{e})$.  where 

\begin{align}
\var{e}&=E(\err{}\err{}^T)  \\
       &=E[(\feature{} - \hat{\feature{}})(\feature{} - \hat{\feature{}})^T]  \\
       &=E[(\boldsymbol{\lambda}\err{}^--(\boldsymbol{I}-\boldsymbol{\lambda}\err{c}))(\boldsymbol{\lambda}\err{}^--(\boldsymbol{I}-\boldsymbol{\lambda}\err{c}))^T] 
\end{align}

$\err{}^-$ represents the prior estimation of $\err{}$, Since $\err{}^-$ and $\err{c}$ are independent of each other, we have $E(\err{c}\err{}^-)=E(\err{}^-)E(\err{c})=0$. Therefore, we have

\begin{align}
    \var{e}&=\boldsymbol{\lambda}E(\err{}^-\err{}^{-T})\boldsymbol{\lambda}^T+(\boldsymbol{I}-\boldsymbol{\lambda})E(\err{c}\err{c}^T)(\boldsymbol{I}-\boldsymbol{\lambda})^T  \\
    &=\boldsymbol{\lambda}\var{}^-\boldsymbol{\lambda}^T+(\boldsymbol{I}-\boldsymbol{\lambda})\var{c}(\boldsymbol{I}-\boldsymbol{\lambda})^T 
\end{align}

In order to estimate $\feature{}$ accurately, we have to minimize $\err{}$, i.e.,
\begin{equation}
    \min\limits_{\lambda}~tr(\var{e})= \min\limits_{\lambda}~[tr(\boldsymbol{\lambda}\var{}^-\boldsymbol{\lambda}^T)-2tr(\boldsymbol{\lambda}\var{c})+tr(\boldsymbol{\lambda}\var{c}\boldsymbol{\lambda}^T)] 
\end{equation}

We need to solve this equation:
\begin{equation}
\frac{\partial\var{e}}{\partial\boldsymbol{\lambda}}=0 
\end{equation}

We have
\begin{equation}
    % \boldsymbol{\lambda} = \frac{\var{c}}{\var{}^- + \var{c}}   % todo, check
    % \boldsymbol{\lambda} = \var{c}(\var{}^-+\var{c})^{-1} 
    \boldsymbol{\lambda} = (\var{}^-\var{c}^{-1}+\boldsymbol{I})^{-1} 
\end{equation}

where
\begin{align}
    \var{}^-&=E(\err{}^-\err{}^{-T})  \\
    &=E[(\feature{}-\hat{\feature{}})(\feature{}-\hat{\feature{}})^T]  \\
    &=E[(\mean{t} + \err{t}-\mean{t}) (\mean{t} + \err{t}-\mean{t})^T]  \\
    &=\var{t} 
\end{align}

Therefore
\begin{equation}
\label{equ:lambda}
    % \boldsymbol{\lambda} = \frac{\var{c}}{\var{t} + \var{c}}  % todo, check
    % \boldsymbol{\lambda} = \var{c}(\var{t}+\var{c})^{-1}
    \boldsymbol{\lambda} = (\var{t}\var{c}^{-1}+\boldsymbol{I})^{-1}
\end{equation}

% 根据公式可以看出，尽管我们无法依据t和c准确估计λ值，但是，我们可以根据patch数的变化判断λ的变化趋势从而辅助我们选择一个合适的超参。当patch数足够多时，闭包引导的图像理解过程能够很好的估计出分布特征，即：ec很小，所以此时λi趋近于0。随着patch数量的降低，闭包估计的图像理解的误差逐渐增加，此时，λ随之增加。
Although the error covariance matrix $\var{t}$ and $\var{c}$ of the observers $O_t$ and $O_c$ cannot be calculated, the relationship between $\boldsymbol{\lambda}$ and the number of patches can still be estimated, which can assist us in choosing the parameter. When the number of patches is large enough, the Closure-Guided image understanding can estimate the image features accurately. At this time, $\var{c}$ is close to $\boldsymbol{0}$. According to \textbf{Equation \ref{equ:lambda}}, $\boldsymbol{\lambda}$ is close to $\boldsymbol{0}$.
As the number of patches decreases, $\var{c}$ gradually increases, and $\boldsymbol{\lambda}$ is close to $\boldsymbol{I}$.

\subsection{The Overview of Our Approach}

\begin{figure}[htbp]
    \centering
    \includegraphics[width=\linewidth]{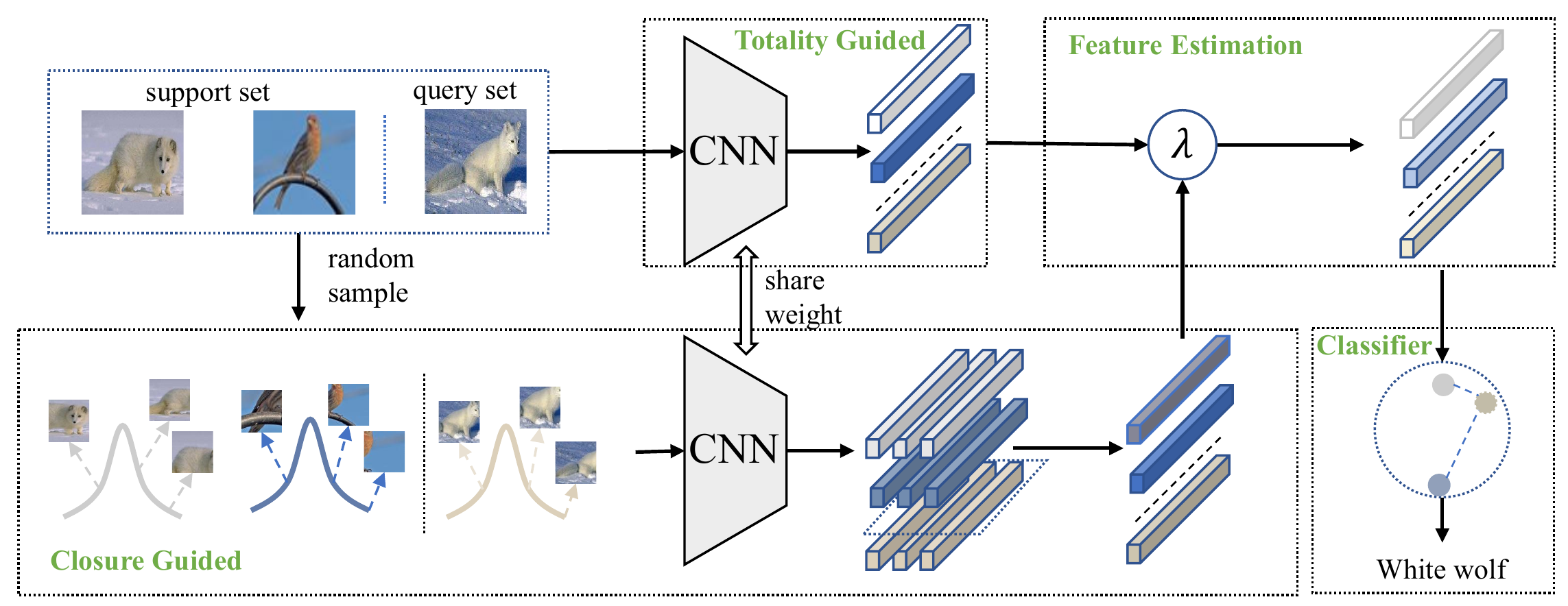}
    % \includesvg[width=\linewidth]{image/method.svg}
    % \caption{This is the overview of our approach. There are two branches in our method. Totality-Guided image understanding module extracts the feature of the whole image. Guided by Closure-Guided image understanding module, image features are estimated from incomplete images. After that, we use feature estimation module to fuse the feature calculated by the Totality-Guided and Closure-Guided image understanding module. Finally, we classify the query image according to the image feature.}
    \caption{There are two branches in our method. Totality-Guided module extracts the feature of the whole image. Guided by Closure-Guided module, image features are estimated from incomplete images. After that, we use feature estimation module to fuse the feature calculated by the Totality-Guided and Closure-Guided module. Finally, we classify the query image according to the image feature.}
    \label{fig:method}
\end{figure}

Our pipeline on 2-way 1-shot task is illustrated in \textbf{Fig. \ref{fig:method}}. Given a few-shot learning task, firstly, we feed the image $x$ into the feature extractor $\extractor{\cdot}$ to extract the features.The prototype guided by the principle of totality $\boldsymbol{P^t_n}$ can be calculated by \textbf{Equation \ref{prototype_t}}. Query features estimated by the principle of totality are the whole image feature extracted by the feature extractor (\textbf{Equation \ref{estimateN1}}).
% can be  calculated by \textbf{Equation \ref{estimateN1}}.

\begin{equation}
    \label{prototype_t}
    \boldsymbol{P^t_n} = \frac{1}{\imageN{}} \sum_{\datapair{i}\in\supportSet_n}\extractor{\image{i}}
\end{equation}

Meanwhile, guided by the law of closure, we randomly crop $M$ patches from each image and feed them into a feature extractor with shared weights to calculate feature $\extractor{p}$. For the convenience of calculating the categories prototypes, as shown in \textbf{Equation \ref{prototype_c}}, we use all patches in the same category to calculate the prototype guided by the law of closure. Query features estimated by the law of closure can be calculated by \textbf{Equation \ref{estimateN2}}.

\begin{equation}
    \label{prototype_c}
     \boldsymbol{P^c_n} = \frac{1}{\imageN{}\times \gpatchN{}} \sum_{\datapair{i}\in\supportSet_n}\sum_{p_i\in\image{i}}\extractor{p_i}
\end{equation}

After calculating the category prototypes \boldsymbol{$P_n^t$} and \boldsymbol{$P_n^c$}, they are fed into the feature estimate module to calculate the category prototype \boldsymbol{$P_n$} (\textbf{Equation \ref{equ:prototype}}). Query features can be re-estimated by \textbf{Equation \ref{equ:imagefea}}. Then the distances between the query feature and the category prototypes are calculated and the query set is classified according to \textbf{Equation \ref{equ:softmax}}.

\begin{equation}
	\label{equ:prototype}
     \boldsymbol{P_n}=\boldsymbol{\lambda} \boldsymbol{P^t_n} + (\boldsymbol{I}-\boldsymbol{\lambda}) \boldsymbol{P^c_n}
\end{equation}

\begin{table}[bp]
    \centering
    \tabcolsep=5mm
    \begin{tabular}{l c c}
        \hline
        method & 5-way 1-shot~($\%$) & 5-way 5-shot~($\%$) \\
          \hline
        PN & $61.59 \pm 0.54$ & $76.75 \pm 0.46$ \\
        PN+GGIU & $64.34 \pm 0.53$ ($\uparrow$ \textbf{2.75}) & $79.49 \pm 0.41$ ($\uparrow$ \textbf{2.74}) \\
        CC & $63.11 \pm 0.74$ & $80.43 \pm 0.31$ \\
        CC+GGIU & $65.72 \pm 0.77$ ($\uparrow$ \textbf{2.61}) & $82.55 \pm 0.29$ ($\uparrow$ \textbf{2.12}) \\
        CL & $63.74 \pm 0.59$ & $79.33 \pm 0.31$ \\
        CL+GGIU & $65.50 \pm 0.45$ ($\uparrow$ \textbf{1.76}) & $80.76 \pm 0.39$ ($\uparrow$ \textbf{1.43}) \\
        CLIP & $88.21 \pm 0.33$ & $97.47 \pm 0.08$ \\
        CLIP+GGIU & $89.31 \pm 0.33$ ($\uparrow$ \textbf{1.10}) & $97.71 \pm 0.06$ ($\uparrow$ \textbf{0.24}) \\
        \hline
    \end{tabular}
    \caption{Results of the performance of different methods on \emph{mini}ImageNet before and after adding GGIU The reported accuracy is 95\% confidence interval.}
    \label{tab:miniImageRes}
\end{table}

\section{Experiment}
\label{sec:exprtiment}

\subsection{Datasets}
We test our method on \emph{mini}ImageNet \cite{DBLP:conf/nips/VinyalsBLKW16} and Caltech-UCSD Birds 200-2011 (CUB200) \cite{wah2011caltech}, which are widely used in few-shot learning.

\textbf{\emph{mini}ImageNet} is a subset of ILSVRC-2012 \cite{DBLP:journals/ijcv/RussakovskyDSKS15}. It contains 60,000 images in 100 categories, with 600 images in each category. Among them, 64 classes are used as the training set, 16 classes as the validation set, and 20 as the testing set.

\textbf{CUB200} contains 11788 images of birds in 200 species, which is widely used for fine-grained classification. Following the previous work \cite{DBLP:journals/corr/abs-2109-04898}, we split the categories into 130, 20, 50 for training, validation and testing.

\subsection{Implementation Details}
Since we propose a test-time feature estimation approach, we need to reproduce the performance of existing methods to validate our approach's effectiveness. Therefore, following Luo \cite{DBLP:conf/nips/LuoWWYXXT21}, we reproduce PN \cite{DBLP:conf/nips/SnellSZ17}, CC \cite{DBLP:conf/cvpr/GidarisK18}, and  CL \cite{DBLP:conf/icmcs/LuoCWPX21}. The backbone we use in this paper is ResNet-12 \cite{DBLP:conf/cvpr/HeZRS16}, which is widely used in few-shot learning. We implement our method using PyTorch and test on an NVIDIA 3090 GPU. Since the authors do not provide a configuration file for the CUB200 dataset, we use the same configuration file as \emph{mini}ImageNet. In the test phase, we randomly sampled five groups of test tasks, and each group of tasks contained 2000 episodes. Then five patches from each image are randomly cropped to rectify for prototypes and features with $\boldsymbol{\lambda} = diag(0.5, 0.5, \ldots, 0.5)$. The size of the patches is range from 0.08 to 1.0 of that of the original image.

\begin{table}[bp]
    \centering
    \tabcolsep=5mm
    \begin{tabular}{l c c}
        \hline
        method & 5-way 1-shot~($\%$) & 5-way 5-shot~($\%$) \\
          \hline
        PN & $76.13 \pm 0.21$  & $88.06 \pm 0.09$  \\
        PN+GGIU & $78.79 \pm 0.24$ ($\uparrow$ \textbf{2.66}) & $89.69 \pm 0.17$ ($\uparrow$ \textbf{1.63}) \\
        CC & $70.57 \pm 0.35$ & $86.65 \pm 0.16$ \\
        CC+GGIU & $72.60 \pm 0.29$ ($\uparrow$ \textbf{2.03}) & $87.90 \pm 0.27$ ($\uparrow$ \textbf{1.25}) \\
        CL & $72.34 \pm 0.48$ & $85.93 \pm 0.25$ \\
        CL+GGIU & $73.64 \pm 0.46$ ($\uparrow$ \textbf{1.30}) & $87.17 \pm 0.25$ ($\uparrow$ \textbf{1.24}) \\
        \hline
    \end{tabular}
    \caption{Results of the performance of different methods on CUB200 before and after adding GGIU. The reported accuracy is 95\% confidence interval.}
    \label{tab:CUBRes}
\end{table}

\begin{table}[t]
    \centering
    \tabcolsep=3mm
    \begin{tabular}{l c c c}
        \toprule [1.5 pt]
        Method & Backbone & 5-way 1-shot~($\%$) & 5-way 5-shot~($\%$) \\
          \midrule [1 pt]
        PN*\cite{DBLP:conf/nips/SnellSZ17} & Conv-4 & $49.42 \pm 0.78$ & \bm{$68.20 \pm 0.66$} \\
        PN\dag & Conv-4 & $50.15 \pm 0.44$ & $65.19 \pm 0.51$ \\
        DC* \cite{DBLP:conf/iclr/YangLX21} & Conv-4 & \bm{$54.62 \pm 0.64$} & - \\
        % CSS(meta-training)* & Conv-4 & $50.85 \pm 0.84$ & $68.08 \pm 0.73$ \\
        Spot and Learn*\cite{chu2019spot} & Conv-4 & $51.03 \pm 0.78$ & $67.96 \pm 0.71$ \\
        \hline
        PN+GGIU\dag & Conv-4 & $52.55 \pm 0.52$  & $67.36 \pm 0.55$   \\
        \midrule [1 pt]
        PN\dag  & ResNet-12 & $61.59 \pm 0.54$ & $76.75 \pm 0.46$ \\
        CC*\cite{DBLP:conf/cvpr/GidarisK18}  & ResNet-12 & $55.45 \pm 0.89$ & $70.13 \pm 0.68$ \\
        CC\dag  & ResNet-12 & $63.11 \pm 0.74$ & $80.43 \pm 0.31$ \\
        % SLA-AG* & ResNet-12 & $62.93 \pm 0.63$ & $79.63 \pm 0.47$ \\
        PN+TRAML*\cite{DBLP:conf/cvpr/Li0LFLW20} & ResNet-12 & $60.31 \pm 0.48$ & $77.94 \pm 0.57$ \\
        PN+CL*\cite{DBLP:conf/icmcs/LuoCWPX21} & ResNet-12 & $59.54 \pm 0.47$ & $74.46 \pm 0.52$ \\
        PN+CL\dag & ResNet-12 & $63.74 \pm 0.59$ & $79.33 \pm 0.31$ \\
        % ConstellationNet*  & ResNet-12 & $64.89 \pm 0.23$ & $79.95 \pm 0.17$ \\
        % Meta-Baseline* & ResNet-12 & $63.17 \pm 0.23$ & $79.26\pm 0.17$ \\
        DC* & ResNet-18 & $61.50 \pm 0.47$ & - \\
        AA*\cite{afrasiyabi2020associative} & ResNet-18 & $58.84 \pm 0.77$ & $80.35 \pm 0.73$ \\
        \hline
        PN+GGIU\dag & ResNet-12 & $64.34 \pm 0.53$  & $79.49 \pm 0.41$   \\
        CC+GGIU\dag & ResNet-12 & \bm{$65.72 \pm 0.77$}  & \bm{$82.55 \pm 0.29$}   \\
        PN+CL+GGIU\dag & ResNet-12 & $65.50 \pm 0.45$  & $80.76 \pm 0.39$   \\
        \midrule [1 pt]
        CLIP\dag & ViT-B/32 & $88.21 \pm 0.33$ & $97.47 \pm 0.08$ \\
        \hline
        CLIP+GGIU\dag & ViT-B/32 & \bm{$89.31 \pm 0.33$} & \bm{$97.71 \pm 0.06$} \\
        \bottomrule [1.5 pt]
    \end{tabular}
    \caption{Results of the performance reported by other methods with the performance of ours on \emph{mini}ImageNet. The reported accuracy is 95\% confidence interval. * represents the results reported by the original paper and \dag \ represents the results that we implement.}
    \label{tab:allresmini}
\end{table}

\begin{table}[!bp]
    \centering
    \tabcolsep=5mm
    \begin{tabular}{l c c}
        \hline
        method & 5-way 1-shot~($\%$) & 5-way 5-shot~($\%$) \\
          \hline
        PN & $40.47 \pm 0.21$  & $56.14 \pm 0.20$  \\
        PN+GGIU & $42.61 \pm 0.49$ ($\uparrow$ \textbf{2.14}) & $58.95 \pm 0.49$ ($\uparrow$ \textbf{2.81}) \\
        CC & $43.56 \pm 0.47$ & $61.51 \pm 0.39$ \\
        CC+GGIU & $45.88 \pm 0.48$ ($\uparrow$ \textbf{2.32}) & $64.77 \pm 0.27$ ($\uparrow$ \textbf{3.26}) \\
        CL & $38.65 \pm 0.44$ & $52.36 \pm 0.35$ \\
        CL+GGIU & $39.87 \pm 0.31$ ($\uparrow$ \textbf{1.22}) & $53.74 \pm 0.21$ ($\uparrow$ \textbf{1.38}) \\
        \hline
    \end{tabular}
    \caption{Results of the performance of different methods trained on \emph{mini}ImageNet and tested on CUB200 before and after adding GGIU. The reported accuracy is 95\% confidence interval.}
    \label{tab:miniCUB}
\end{table}

\subsection{Experiment Results}

\subsubsection{Results on in-domain data}

\textbf{Table~\ref{tab:miniImageRes}} shows our performance on \emph{mini}ImageNet. Our method can effectively improve the performance based on existing methods for different tasks. we improve the performance on PN, CC, and CL by 2.75\%, 2.61\%, and 1.76\%, separately.
% On PN, we improve it by 2.75\%.

We also test the performance on CLIP\cite{DBLP:conf/icml/RadfordKHRGASAM21} to explore the performance of our method on the model pre-trained on large-scale data. We use the ViT-B/32 model published by OpenAI as a feature extractor and use PN for classification. Surprising, on such a high baseline, our method can still improve the 1-shot task by $1.10\%$ and the 5-shot by $0.23\%$. The experiment results also illustrate that even with such large-scale training data, \textbf{GGIU} can still estimate more accurate image features.

Similarly, to test the effectiveness of our method on fine-grained classification, we also test the performance of our method on CUB200. As shown in \textbf{Table~\ref{tab:CUBRes}}, our method also significantly improves the performance based on existing methods of fine-grained classification.

In addition, we compare the performance of our method with existing methods in \textbf{Table~\ref{tab:allresmini}}.

\subsubsection{Results on cross-domain data}

To validate the effectiveness and robustness of our approach, we conduct experiments on cross-domain tasks. We test the cross-domain performance on \emph{mini}ImageNet and CUB200: \textbf{Table~\ref{tab:miniCUB}} shows the model trained on \emph{mini}ImageNet and tested on CUB200; \textbf{Table~\ref{tab:CUBmini}} shows the results of the model trained on CUB200 and tested on \emph{mini}ImageNet. It can be seen that our method has adequate performance improvement on the cross-domain task of few-shot learning.

\begin{table}[tbp]
    \centering
    \tabcolsep=5mm
    \begin{tabular}{l c c}
        \hline
        method & 5-way 1-shot~($\%$) & 5-way 5-shot~($\%$) \\
          \hline
        PN & $40.24 \pm 0.35$  & $55.47 \pm 0.47$  \\
        PN+GGIU & $43.17 \pm 0.57$ ($\uparrow$ \textbf{2.93}) & $58.12 \pm 0.49$ ($\uparrow$ \textbf{2.65}) \\
        CC & $43.54 \pm 0.52$ & $60.40 \pm 0.39$ \\
        CC+GGIU & $44.27 \pm 0.42$ ($\uparrow$ \textbf{0.73}) & $60.94 \pm 0.42$ ($\uparrow $ \textbf{0.54}) \\
        CL & $44.47 \pm 0.56$ & $61.84 \pm 0.44$ \\
        CL+GGIU & $46.42 \pm 0.68$ ($\uparrow$ \textbf{1.95}) & $63.89 \pm 0.45$ ($\uparrow$ \textbf{2.05}) \\
        \hline
    \end{tabular}
    \caption{Results of the performance of different methods trained on CUB200 and tested on miniImageNet. The reported accuracy is 95\% confidence interval.}
    \label{tab:CUBmini}
\end{table}

\subsection{Result Analysis}
\subsubsection{Ablation Study}
This section performs ablation experiments on \emph{mini}ImageNet to explore the performance impact of feature rectification on the support set and query set.

As shown in \textbf{Table~\ref{tab:ablation}}, compared with only using \textbf{GGIU} to estimate query features, the performance of estimation support features is higher. For the 5-way 5-shot task, when \textbf{GGIU} is used to estimate query features, the performance is improved more. This can be explained as follows: The more support set samples, the more accurate the representation of the category prototype is. At this time, the accuracy of query features is the bottleneck of classification performance. Similarly, when the support set samples are few, the category prototypes calculated by the support set are inaccurate. At this moment, accurate category prototypes can significantly improve classification performance.

\subsubsection{The Influence of The Fusion Parameters}

In our method, the fusion parameter $\boldsymbol{\lambda}=diag(\lambda_1, \lambda_2, \ldots, \lambda_D)$ is essential. This section explores the influence of $\boldsymbol{\lambda}$. We conduct three sets of experiments on PN for the different number of patches, 1, 5, and 10, respectively. As shown in \textbf{Fig. \ref{fig:lambdaacc}}, when $\lambda_i=1$, image features $\hat{\feature{}}$ are solely estimated by totality-guided image understanding (\textbf{Equation.\ref{equ:imagefea}}). The performance at this point is the baseline performance. As $\lambda_i$ decreases, the influence of closure-guided image understanding increases, and the estimated image feature $\hat{\feature{}}$ can represent the image better. So the performance gradually improves until an equilibrium point is reached. After the highest performance, continuing to decrease $\lambda_i$ leads to a gradual decrease in model performance.
It is worth noting that the performance will improve with a large enough number of patches even if $\lambda_i = 0$. It can be concluded that when the number of patches is large enough, closure-guided image understanding can well estimate image features. It can also be seen that the value of $\lambda$ corresponding to the highest performance decreases as the number of patches increases. The more patches, the more accurate the estimation guided by the law of closure. What's more, when the number of patches is 1, 5, and 10, the best $\lambda$ equals 0.7, 0.5, and 0.4, respectively. The above results are consistent with the deduction in \textbf{Section.\ref{sec:lambda}}.

\begin{table}[tp]
    \centering
    \tabcolsep=3mm
    \begin{tabular}{c c c c}
        \hline
       support & query & 5-way 1-shot~($\%$) & 5-way 5-shot~($\%$) \\
          \hline
         &  & 61.59  & 76.75  \\
        \checkmark &  & 62.80  & 77.01  \\
         & \checkmark & 62.50  & 77.90  \\
        \checkmark & \checkmark & 64.34  & 79.49 \\
        \hline
    \end{tabular}
    \caption{The results of ablation experiments. support with $\checkmark$ means adding GGIU while calculating prototypes; query with $\checkmark$ means adding GGIU while computing the feature of query images.}
    \label{tab:ablation}
\end{table}

\begin{figure}[bp]
\begin{minipage}[t]{0.48\columnwidth}
\centering
% \begin{figure}[h] %插入图片
% 		\centering %图片居中
% 		\resizebox{0.5\columnwidth}{!}{  %用于修改图片大小
			\begin{tikzpicture} %tikz图片
			\scalefont{1} %设置字体大小
			\begin{axis}[
			sharp plot, %控制线的风格
% 			title=line chart,%图像标题
			xmode=normal,% 控制坐标轴为线性
%		ymode=log,% 控制坐标轴为对数
			xlabel=\bm{$\lambda$}, %x坐标名
			ylabel=acc(\%), %y坐标名
			width=\linewidth,  %设置长和宽
			xmin=-0.03,xmax=1.03,  % 设置x坐标范围
			ymin=54, ymax=66,  % 设置y坐标范围
			xtick={}, %指定x轴刻度值。如果为空，则自动设置刻度线。即分割坐标轴
			ytick={54, 56, 58, 60, 62, 64, 66}, %指定y轴刻度值。如果为空，则自动设置刻度线。即分割坐标轴
			xlabel near ticks, % 设置x坐标名位置靠近折线图
			ylabel near ticks, % 设置y坐标名位置靠近折线图
			ymajorgrids=true, % 启用/禁用 [公式] 轴上刻度线位置上的网格线
			grid style=dashed, % 设置网格线格式
% 			legend style={at={(0.68,0.06)},anchor=south}, % 设置标签位置
%			legend columns=3, %设置标签列数
            % legend( font=1
			legend pos=south east, % 设置折线对应标签的位置
			legend style={nodes={scale=0.6, transform shape}},  % 设置折线标签的格式
			]
			
			\addplot+[semithick,mark=x,mark options={scale=0.6}, color=color0] plot coordinates { 
			    (0.0, 61.58)
			    (0.1, 61.58)
			    (0.2, 61.58)
			    (0.3, 61.58)
			    (0.4, 61.58)
			    (0.5, 61.58)
			    (0.6, 61.58)
			    (0.7, 61.58)
			    (0.8, 61.58)
			    (0.9, 61.58)
			    (1.0, 61.58)
			};
			\addlegendentry{baseline}%第一条线标签
			
			%画第一条线，semithick设置线的粗细为0.6pt，mark是折线标示形状，options是mark形状的大小 ， olive!50!white是颜色，coordinates中包含要绘制的点的坐标
			\addplot+[semithick,mark=x,mark options={scale=0.6}, color=color1] plot coordinates { 
			    (0.0, 55.18)
			    (0.1, 57.27)
			    (0.2, 59.05)
			    (0.3, 60.61)
			    (0.4, 61.77)
			    (0.5, 62.59)
			    (0.6, 62.97)
			    (0.7, 62.99)
			    (0.8, 62.75)
			    (0.9, 62.23)
			    (1.0, 61.58)
			};
			\addlegendentry{number of patches = 1}%第一条线标签
			
			%画第二条线
			\addplot+[semithick,mark options={scale=0.3}, color=color2] plot coordinates {
				(0.0, 62.03)
			    (0.1, 62.86)
			    (0.2, 63.51)
			    (0.3, 63.97)
			    (0.4, 64.19)
			    (0.5, 64.23)
			    (0.6, 64.07)
			    (0.7, 63.72)
			    (0.8, 63.11)
			    (0.9, 62.38)
			    (1.0, 61.58)
			};
			\addlegendentry{number of patches = 5} %第二条线标签
			
			%画第三条线
			\addplot+[semithick,mark options={scale=0.3}, color=color3] plot coordinates {
				(0.0, 63.01)
			    (0.1, 63.65)
			    (0.2, 64.12)
			    (0.3, 64.42)
			    (0.4, 64.55)
			    (0.5, 64.47)
			    (0.6, 64.24)
			    (0.7, 63.80)
			    (0.8, 63.18)
			    (0.9, 62.42)
			    (1.0, 61.58)
			};
			\addlegendentry{number of patches = 10} %第三条线标签

			\end{axis}
			\end{tikzpicture}
% 		}
		\caption{Relationship between $\boldsymbol{\lambda}$ and performance} % 设置caption
		\label{fig:lambdaacc}  % 设置用于reference的label
% 	\end{figure}
\end{minipage}%
\hfill
\begin{minipage}[t]{0.48\columnwidth}
\centering
\begin{tikzpicture} %tikz图片
			\scalefont{1} %设置字体大小
			\begin{axis}[
			sharp plot, %控制线的风格
% 			title=line chart,%图像标题
			xmode=normal,% 控制坐标轴为线性
%		ymode=log,% 控制坐标轴为对数
			xlabel=number of patches, %x坐标名
			ylabel=acc(\%), %y坐标名
			width=\linewidth,   %设置长和宽
			xmin=-1,xmax=22,  % 设置x坐标范围
			ymin=61, ymax=65,  % 设置y坐标范围
			xtick={}, %指定x轴刻度值。如果为空，则自动设置刻度线。即分割坐标轴
			ytick={61, 62, 63, 64}, %指定y轴刻度值。如果为空，则自动设置刻度线。即分割坐标轴
			xlabel near ticks, % 设置x坐标名位置靠近折线图
			ylabel near ticks, % 设置y坐标名位置靠近折线图
			ymajorgrids=true, % 启用/禁用 [公式] 轴上刻度线位置上的网格线
			grid style=dashed, % 设置网格线格式
% 			legend style={at={(0.68,0.06)},anchor=south}, % 设置标签位置
%			legend columns=3, %设置标签列数
			legend pos=south east, % 设置折线对应标签的位置
%			legend style={nodes={scale=0.6, transform shape}},  % 设置折线标签的格式
			]
			[61.58, 62.59, 63.54, 63.93, 64.13, 64.22, 64.29, 64.37, 64.40, 64.46, 64.47, 
			64.50, 64.53, 64.54, 64.56, 64.54, 64.59, 64.56, 64.55, 64.58, 64.59, 64.64]
			\addplot+[semithick,mark=x,mark options={scale=0.6}, color=color1] plot coordinates { 
			    (0, 61.58)
			    (1, 62.59)
			    (2, 63.54)
			    (3, 63.93)
			    (4, 64.13)
			    (5, 64.22)
			    (6, 64.29)
			    (7, 64.37)
			    (8, 64.40)
			    (9, 64.46)
			    (10, 64.47)
			    (11, 64.50)
			    (12, 64.53)
			    (13, 64.54)
			    (14, 64.56)
			    (15, 64.54)
			    (16, 64.59)
			    (17, 64.56)
			    (18, 64.55)
			    (19, 64.58)
			    (20, 64.59)
			    (20, 64.64)
			};
% 			\addlegendentry{baseline}%第一条线标签

			\end{axis}
			\end{tikzpicture}
% 		}
		\caption{The influence of the number of patches on performance} % 设置caption
		\label{fig:numacc}  % 设置用于reference的label
\end{minipage}
\end{figure}

\subsubsection{The Influence of The Number of Patches}
The number of patches is a very important hyper-parameter, and the inference efficiency might be low if this number is too big. This section explores how the number of patches influences the performance. As shown in \textbf{Fig. \ref{fig:numacc}}, on \emph{mini}ImageNet, we perform a 5-way 1-shot experiment on PN with $\lambda_i=0.5$. Suppose only one patch is cropped to estimate the closure feature, in which case, it can be seen that it will also improve the performance substantially. However, as the number increases, the rate of increase in accuracy gradually slows down, which demonstrates that too many patches might lead to a marginal effect on the correction of the distribution, especially when the patches almost cover the whole image. Therefore, too many patches do not significantly improve the model performance.

\subsubsection{The Relationship between intra-class variations and \bm{$\lambda$}}

\begin{figure}[tp]
\centering
\includegraphics[width=0.6\linewidth]{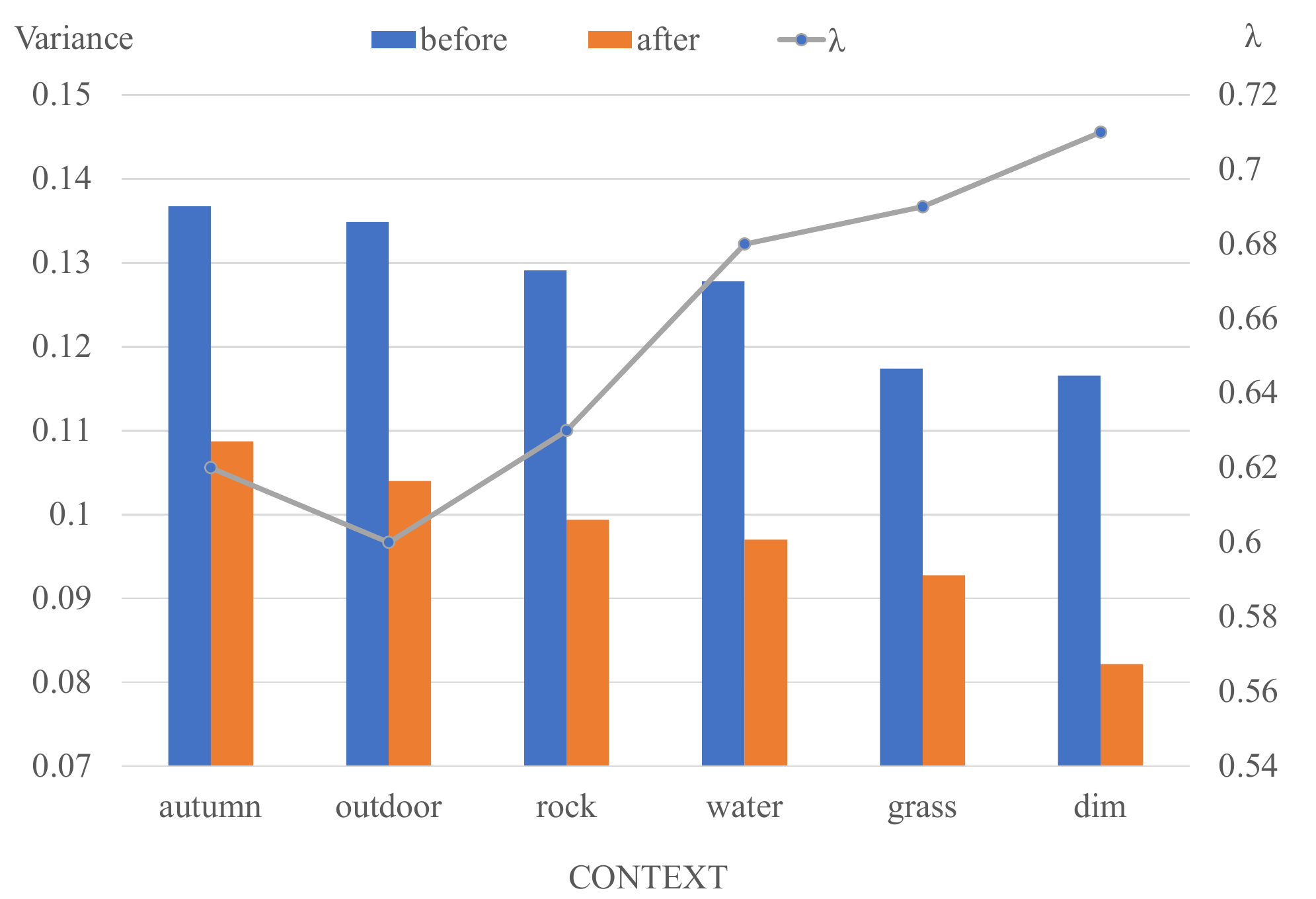}
\caption{The relationship between intra-class variations and \bm{$\lambda$}}
\label{fig:lam_var}
\end{figure}

In order to analyse the relationship between the optimal \bm{$\lambda$} and intra-class variations, we conducted experiments on the NICO\cite{zhang2022nico}, which contains images labeled with category and context. We searched for the optimal \bm{$\lambda$} in different contexts and calculated the intra-class variance in each context before and after using our method. As shown in \textbf{Fig. \ref{fig:lam_var}}, with the decrease of the intra-class variance, the optimal \bm{$\lambda$} shows an increasing trend. Moreover, our method can significantly reduce the intra-class variance.

\section{Conclusion}
\label{sec:conclusion}

In this paper, we reformulate image features from the perspective of multivariate Gaussian distributions. We introduce Gestalt psychology into the process of image understanding to estimate more accurate image features. The Gestalt-guided image understanding consists of two modules: Totality-guided image understanding and Closure-guidied image understanding. Then we fed the features obtained from the above two modules into the feature estimation module and estimate image features accuractely. We conduct many experiments on \emph{mini}ImageNet and CUB200 for coarse-grained, fine-grained, and cross-domain few-shot image classification. The results demonstrate the effectiveness of \textbf{GGIU}. Moreover, \textbf{GGIU} even improved the performance based on CLIP. Finally, we analyze the influence of different hyper-parameters, and the results accord with our theoretical analysis.

\subsubsection{\ackname} This work was supported by the National Natural Science Foundation of China (No. U20B2062 and No. 62172036), the Fundamental Research Funds for the Central Universities (No. FRF-TP-20-064A1Z), the key Laboratory of Opto-Electronic Information Processing, CAS (No. JGA202004027), and the R$\&$D Program of CAAC Key Laboratory of Flight Techniques and Flight Safety (NO. FZ2021ZZ05).

\bibliographystyle{splncs04}
\bibliography{mybibliography}

\begin{thebibliography}{10}
\providecommand{\url}[1]{\texttt{#1}}
\providecommand{\urlprefix}{URL }
\providecommand{\doi}[1]{https://doi.org/#1}

\bibitem{DBLP:conf/icml/RadfordKHRGASAM21}
Radford, A., Kim, J.W., Hallacy, C., Ramesh, A., Goh, G., Agarwal, S., Sastry,
  G., Askell, A., Mishkin, P., Clark, J., Krueger, G., Sutskever, I.: Learning
  transferable visual models from natural language supervision. In: Proceedings
  of the 38th International Conference on Machine Learning, {ICML} 2021, 18-24
  July 2021, Virtual Event. Proceedings of Machine Learning Research, vol.~139,
  pp. 8748--8763 (2021)

\bibitem{DBLP:journals/pami/Fei-FeiFP06}
Fei{-}Fei, L., Fergus, R., Perona, P.: One-shot learning of object categories.
  {IEEE} Trans. Pattern Anal. Mach. Intell.  \textbf{28}(4),  594--611 (2006).
  \doi{10.1109/TPAMI.2006.79}

\bibitem{DBLP:conf/nips/VinyalsBLKW16}
Vinyals, O., Blundell, C., Lillicrap, T., Kavukcuoglu, K., Wierstra, D.:
  Matching networks for one shot learning. In: Advances in Neural Information
  Processing Systems 29: Annual Conference on Neural Information Processing
  Systems 2016, December 5-10, 2016, Barcelona, Spain. pp. 3630--3638 (2016)

\bibitem{DBLP:conf/nips/SnellSZ17}
Snell, J., Swersky, K., Zemel, R.S.: Prototypical networks for few-shot
  learning. In: Advances in Neural Information Processing Systems 30: Annual
  Conference on Neural Information Processing Systems 2017, December 4-9, 2017,
  Long Beach, CA, {USA}. pp. 4077--4087 (2017)

\bibitem{DBLP:conf/icml/FinnAL17}
Finn, C., Abbeel, P., Levine, S.: Model-agnostic meta-learning for fast
  adaptation of deep networks. In: Proceedings of the 34th International
  Conference on Machine Learning, {ICML} 2017, Sydney, NSW, Australia, 6-11
  August 2017. Proceedings of Machine Learning Research, vol.~70, pp.
  1126--1135 (2017)

\bibitem{DBLP:conf/iccv/Chen00D021}
Chen, Y., Liu, Z., Xu, H., Darrell, T., Wang, X.: Meta-baseline: Exploring
  simple meta-learning for few-shot learning. In: 2021 {IEEE/CVF} International
  Conference on Computer Vision, {ICCV} 2021, Montreal, QC, Canada, October
  10-17, 2021. pp. 9042--9051 (2021). \doi{10.1109/ICCV48922.2021.00893}

\bibitem{DBLP:conf/cvpr/ZhangCLS20}
Zhang, C., Cai, Y., Lin, G., Shen, C.: Deepemd: Few-shot image classification
  with differentiable earth mover's distance and structured classifiers. In:
  2020 {IEEE/CVF} Conference on Computer Vision and Pattern Recognition, {CVPR}
  2020, Seattle, WA, USA, June 13-19, 2020. pp. 12200--12210 (2020).
  \doi{10.1109/CVPR42600.2020.01222}

\bibitem{DBLP:journals/corr/abs-2004-05439}
Hospedales, T., Antoniou, A., Micaelli, P., Storkey, A.: Meta-learning in
  neural networks: A survey. IEEE transactions on pattern analysis and machine
  intelligence  \textbf{44}(9),  5149--5169 (2021)

\bibitem{DBLP:books/sp/98/ThrunP98}
Thrun, S., Pratt, L.: Learning to learn: Introduction and overview. In:
  Learning to learn, pp. 3--17. Springer (1998)

\bibitem{koch2015siamese}
Koch, G., Zemel, R., Salakhutdinov, R., et~al.: Siamese neural networks for
  one-shot image recognition. In: ICML deep learning workshop. vol.~2, p.~0.
  Lille (2015)

\bibitem{DBLP:conf/iclr/SatorrasE18}
Satorras, V.G., Estrach, J.B.: Few-shot learning with graph neural networks.
  In: 6th International Conference on Learning Representations, {ICLR} 2018,
  Vancouver, BC, Canada, April 30 - May 3, 2018, Conference Track Proceedings
  (2018)

\bibitem{DBLP:conf/nips/LuoWWYXXT21}
Luo, X., Wei, L., Wen, L., Yang, J., Xie, L., Xu, Z., Tian, Q.: Rectifying the
  shortcut learning of background for few-shot learning. Advances in Neural
  Information Processing Systems  \textbf{34},  13073--13085 (2021)

\bibitem{DBLP:conf/iclr/YangLX21}
Yang, S., Liu, L., Xu, M.: Free lunch for few-shot learning: Distribution
  calibration. In: 9th International Conference on Learning Representations,
  {ICLR} 2021, Virtual Event, Austria, May 3-7, 2021 (2021)

\bibitem{DBLP:conf/ijcai/TangTZ021}
Tang, X., Teng, Z., Zhang, B., Fan, J.: Self-supervised network evolution for
  few-shot classification. In: Zhou, Z. (ed.) Proceedings of the Thirtieth
  International Joint Conference on Artificial Intelligence, {IJCAI} 2021,
  Virtual Event / Montreal, Canada, 19-27 August 2021. pp. 3045--3051 (2021).
  \doi{10.24963/ijcai.2021/419}

\bibitem{DBLP:conf/ijcai/AnXZZ21}
An, Y., Xue, H., Zhao, X., Zhang, L.: Conditional self-supervised learning for
  few-shot classification. In: Zhou, Z. (ed.) Proceedings of the Thirtieth
  International Joint Conference on Artificial Intelligence, {IJCAI} 2021,
  Virtual Event / Montreal, Canada, 19-27 August 2021. pp. 2140--2146 (2021).
  \doi{10.24963/ijcai.2021/295}

\bibitem{li2021learning}
Li, J., Wang, Z., Hu, X.: Learning intact features by erasing-inpainting for
  few-shot classification. In: Proceedings of the AAAI Conference on Artificial
  Intelligence. vol.~35, pp. 8401--8409 (2021)

\bibitem{afrasiyabi2020associative}
Afrasiyabi, A., Lalonde, J.F., Gagn{\'e}, C.: Associative alignment for
  few-shot image classification. In: European Conference on Computer Vision.
  pp. 18--35. Springer (2020)

\bibitem{wagemans2012century}
Wagemans, J., Feldman, J., Gepshtein, S., Kimchi, R., Pomerantz, J.R., Van~der
  Helm, P.A., Van~Leeuwen, C.: A century of gestalt psychology in visual
  perception: Ii. conceptual and theoretical foundations. Psychological
  bulletin  \textbf{138}(6), ~1218 (2012)

\bibitem{henle1972selected}
Henle, M.: The selected papers of wolfgang k{\"o}hler. Philosophy and
  Phenomenological Research  \textbf{33}(2) (1972)

\bibitem{hamlyn2017psychology}
Hamlyn, D.W.: The psychology of perception: A philosophical examination of
  Gestalt theory and derivative theories of perception. Routledge (2017)

\bibitem{brennan2017history}
Brennan, J.F., Houde, K.A.: History and systems of psychology. Cambridge
  University Press (2017)

\bibitem{stevenson2012emergence}
Stevenson, H.: Emergence: The gestalt approach to change. Unleashing executive
  and orzanizational potential. Retrieved  \textbf{7} (2012)

\bibitem{wah2011caltech}
Wah, C., Branson, S., Welinder, P., Perona, P., Belongie, S.: The caltech-ucsd
  birds-200-2011 dataset  (2011)

\bibitem{DBLP:journals/ijcv/RussakovskyDSKS15}
Russakovsky, O., Deng, J., Su, H., Krause, J., Satheesh, S., Ma, S., Huang, Z.,
  Karpathy, A., Khosla, A., Bernstein, M.S., Berg, A.C., Fei{-}Fei, L.:
  Imagenet large scale visual recognition challenge. Int. J. Comput. Vis.
  \textbf{115}(3),  211--252 (2015). \doi{10.1007/s11263-015-0816-y},
  \url{https://doi.org/10.1007/s11263-015-0816-y}

\bibitem{DBLP:journals/corr/abs-2109-04898}
Li, W., Dong, C., Tian, P., Qin, T., Yang, X., Wang, Z., Huo, J., Shi, Y.,
  Wang, L., Gao, Y., et~al.: Libfewshot: A comprehensive library for few-shot
  learning. arXiv preprint arXiv:2109.04898  (2021)

\bibitem{DBLP:conf/cvpr/GidarisK18}
Gidaris, S., Komodakis, N.: Dynamic few-shot visual learning without
  forgetting. In: 2018 {IEEE} Conference on Computer Vision and Pattern
  Recognition, {CVPR} 2018, Salt Lake City, UT, USA, June 18-22, 2018. pp.
  4367--4375 (2018). \doi{10.1109/CVPR.2018.00459}

\bibitem{DBLP:conf/icmcs/LuoCWPX21}
Luo, X., Chen, Y., Wen, L., Pan, L., Xu, Z.: Boosting few-shot classification
  with view-learnable contrastive learning. In: 2021 {IEEE} International
  Conference on Multimedia and Expo, {ICME} 2021, Shenzhen, China, July 5-9,
  2021. pp.~1--6 (2021). \doi{10.1109/ICME51207.2021.9428444}

\bibitem{DBLP:conf/cvpr/HeZRS16}
He, K., Zhang, X., Ren, S., Sun, J.: Deep residual learning for image
  recognition. In: 2016 {IEEE} Conference on Computer Vision and Pattern
  Recognition, {CVPR} 2016, Las Vegas, NV, USA, June 27-30, 2016. pp. 770--778
  (2016). \doi{10.1109/CVPR.2016.90}

\bibitem{chu2019spot}
Chu, W.H., Li, Y.J., Chang, J.C., Wang, Y.C.F.: Spot and learn: A
  maximum-entropy patch sampler for few-shot image classification. In:
  Proceedings of the IEEE/CVF conference on computer vision and pattern
  recognition. pp. 6251--6260 (2019)

\bibitem{DBLP:conf/cvpr/Li0LFLW20}
Li, A., Huang, W., Lan, X., Feng, J., Li, Z., Wang, L.: Boosting few-shot
  learning with adaptive margin loss. In: 2020 {IEEE/CVF} Conference on
  Computer Vision and Pattern Recognition, {CVPR} 2020, Seattle, WA, USA, June
  13-19, 2020. pp. 12573--12581 (2020). \doi{10.1109/CVPR42600.2020.01259}

\bibitem{zhang2022nico}
Zhang, X., Zhou, L., Xu, R., Cui, P., Shen, Z., Liu, H.: Nico++: Towards better
  benchmarking for domain generalization. arXiv preprint arXiv:2204.08040
  (2022)

\end{thebibliography}
\end{document}